\documentclass[11pt]{article}

\usepackage[margin=2.5cm]{geometry}
\usepackage{times}
\usepackage[utf8]{inputenc}
\usepackage[T1]{fontenc}
\usepackage{amsmath,amssymb}
\usepackage{graphicx}
\usepackage{float}
\graphicspath{{figures/}}
\usepackage[hidelinks]{hyperref}
\usepackage[numbers,sort&compress]{natbib}
\usepackage{xcolor}
\usepackage{booktabs}
\usepackage{multirow}
\usepackage{lineno}
\usepackage{setspace}
\usepackage{caption}
\usepackage{subcaption}
\usepackage{authblk}
\usepackage{url}
\usepackage{doi}
\usepackage{microtype}
\usepackage{fancyhdr}
\makeatletter
\g@addto@macro{\UrlBreaks}{\UrlOrds}
\makeatother

\fancypagestyle{firstpage}{%
  \fancyhf{}%
  \fancyfoot[L]{\small *Correspondence to: Dr James K Ruffle, Institute of Neurology, UCL, London WC1N 3BG, UK. Email: j.ruffle@ucl.ac.uk}%
}

\renewenvironment{abstract}{\section*{Abstract}\noindent}{}

\title{\textbf{Fairboard: a quantitative framework for equity assessment of healthcare models}}

\author[1,*]{James K. Ruffle}
\author[2]{Samia Mohinta}
\author[3]{Chris Foulon}
\author[1]{Mohamad Zeina}
\author[1]{Zicheng Wang}
\author[1]{\\Sebastian Brandner}
\author[1]{Harpreet Hyare}
\author[1]{Parashkev Nachev}
\affil[1]{Queen Square Institute of Neurology, University College London, London, UK}
\affil[2]{Dept. of Physiology, Development and Neuroscience, University of Cambridge, UK}
\affil[3]{Laboratoire Bordelais de Recherche en Informatique (LABRI), Universit\'{e} de Bordeaux, France}

\date{}

\begin{document}

\maketitle
\thispagestyle{firstpage}
\vspace{-2em}

\begin{abstract}Despite there now being more than 1,000 FDA-authorised AI medical devices, formal equity assessments---whether model performance is uniform across patient subgroups---are rare. Here, we evaluate the equity of 18 open-source brain tumour segmentation models across 648 glioma patients from two independent datasets ($n$ = 11,664 model inferences) along distinct univariate, Bayesian multivariate, spatial, and representational dimensions. We find that patient identity consistently explains more performance variance than model choice, with clinical factors, including molecular diagnosis, tumour grade, and extent of resection, predicting segmentation accuracy more strongly than model architecture. A voxel-wise spatial meta-analysis identifies neuroanatomically localised biases that are compartment-specific yet often consistent across models. Within a high-dimensional latent space of lesion masks and clinic-demographic features, model performance clusters significantly, indicating that the patient feature space contains axes of algorithmic vulnerability. Although newer models tend toward greater equity, none provide a formal fairness guarantee. Lastly, we release Fairboard, an open-source, no-code dashboard that lowers barriers to equitable model monitoring in medical imaging.
\end{abstract}

\vfill

\newpage

\section*{Main}

\begin{sloppypar}
Deep learning has transformed medical image analysis, achieving expert-level performance across diverse clinical tasks from chest radiograph interpretation to histopathological diagnosis~\citep{litjens2017survey, topol2019highperformance, aggarwal2021diagnostic}. Brain tumour segmentation, the automated delineation of tumour sub-regions from multimodal magnetic resonance imaging (MRI), exemplifies this progress, where global series such as the Brain Tumor Segmentation (BraTS) challenge ~\citep{menze2015brats, bakas2017advancing, bakas2018brats, baid2021rsna} have fostered development of increasingly sophisticated architectures, from encoder--decoder convolutional networks~\citep{isensee2021nnunet, myronenko2019segresnet, ruffle2023segmentation, ruffle2025enhancement} to vision transformers~\citep{hatamizadeh2022swinunetr}, many of which are publicly available and integrated into clinical research pipelines via frameworks such as the BraTS Toolkit~\citep{kofler2020bratstoolkit}. The priority in these developments has been to optimise these models to \emph{overall} performance at a population level, with notably little -- if any -- attention given to an assessment of whether such models perform \emph{equitably} across a diverse patient cohort. Equity, as defined by the World Health Organization (WHO), denotes the absence of unfair, avoidable, or remediable differences among groups of people, whether those groups are defined socially, economically, demographically, geographically, or by other dimensions of inequality~\citep{who2025equity}.
\end{sloppypar}

The issue is not unique to neuro-oncology, but indeed spans much of current healthcare research and development. Swathes of artificial intelligence (AI) research are published daily, though most report a low-dimensional (often singular) metric of cohort performance, that can provide no assurance of equity to the population it is intended to benefit. This gap persists even at the regulatory level: the U.S. Food and Drug Administration (FDA) has now authorised over 1,000 AI/ML-enabled medical devices, the majority in radiology~\citep{muehlematter2021fda, zhang2024fda, milam2023fda}, yet a recent analysis of 691 FDA-cleared devices found that 95.5\% lacked any demographic reporting about the patient populations used in their evaluation~\citep{lin2025fda}. The broader algorithmic fairness literature~\citep{mehrabi2022survey, chouldechova2017fair, xu2022algorithmic} has documented how machine learning systems can encode and amplify societal biases, from facial recognition systems exhibiting accuracy disparities across gender and skin type~\citep{buolamwini2018gender} to clinical risk scores systematically underestimating the healthcare needs of Black patients~\citep{obermeyer2019dissecting}. Medical imaging is no exception: deep learning classifiers trained on chest radiographs exhibit underdiagnosis bias in underserved populations~\citep{seyyedkalantari2021underdiagnosis}, gender imbalances in training data produce biased computer-aided diagnosis systems~\citep{larrazabal2020gender}, algorithmic models can encode protected demographic characteristics even when these are not provided as inputs~\citep{glocker2023algorithmic, ruffle2024computational}, representational biases in model calibration can systematically disadvantage minority groups~\citep{carruthers2022representational}, and fairness corrections effective within training distributions can fail to generalise to external populations~\citep{yang2024limits}. These findings have prompted calls for routine fairness evaluation in medical AI development~\citep{chen2021ethical, mccradden2020ethical, parikh2019addressing}.

Existing fairness assessments in medical imaging have largely been limited to two low-dimensional analytical comparators: group-level performance comparisons and aggregate disparity metrics~\citep{puyol2022fairness, ricci2022addressing, petersen2023feature, xu2024addressing}. However, there are several additional dimensions that are essential but remain underexplored. First, multivariate \emph{cohort equity}, where model errors are fitted as the target variable to a predictive model with any patient features as its input. An equitable model would show no predictive signal in such a fit, whereas an inequitable one would yield features significantly predictive of its performance. Secondly \emph{spatial equity}, an assessment of where model errors are anatomically localised, such as in vision models for brain tumour segmentation. A spatially equitable model should convey no locational performance advantage, yet an inequitable one would delineate anatomical regions in which a stronger or weaker performance occurs. Thirdly, \emph{representational equity}, the assessment of whether a latent space representation of high-dimensional patient features are predictive of variations in model performance. An equitable model would not illustrate a relation between the patient feature latent space and model performance, whereas an inequitable one would identify latent space coordinates that convey a favourable or unfavourable advantage to the model. Lastly, composite equity benchmarks that jointly rank model performance and fairness remain, to our knowledge, absent from the literature, yet could be desirable to enable standardised comparison of models along both accuracy and equity dimensions simultaneously.

\begin{figure}[H]
\centering
\includegraphics[width=\textwidth]{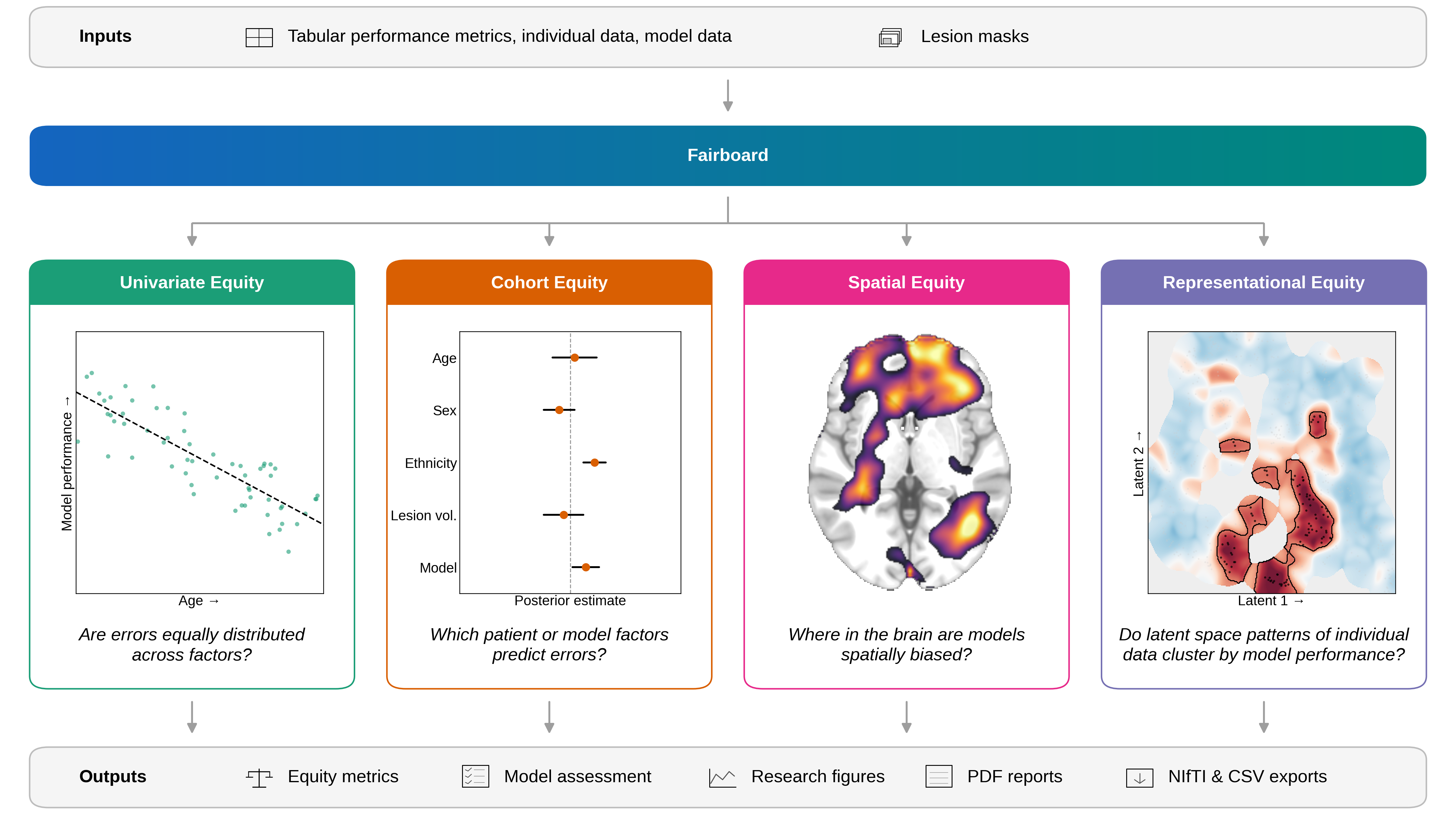}
\caption*{\textbf{Fig.~1. Fairboard: an equitable model monitoring dashboard.}
Schematic overview of the Fairboard platform architecture. \textbf{Top:} Input data types accepted by the dashboard, including tabular performance metrics (individual-level and model-level data) and lesion masks (in zipped NIfTI format). \textbf{Centre:} The Fairboard processing hub, comprising four modules: (i) \emph{Univariate Equity}---statistics and plotting tools to assess whether model errors are equally distributed across univariate individual factors; (ii) \emph{Cohort Equity}---multivariable Bayesian and frequentist regressions to identify which patient or model factors predict model errors; (iii) \emph{Spatial Equity}---brain voxel-wise generalised linear models (GLMs) to test if and where in the brain models are spatially biased; and (iv) \emph{Representational Equity}---dimensionality reduction (e.g.\ UMAP) models with latent-space visualisations and GLM evaluations to test whether patterns of model performance cluster by characteristics in a representational latent space. \textbf{Bottom:} Output deliverables include equity metrics, model assessment reports, publication-ready figures, PDF reports, and NIfTI/CSV exports. GLM, generalised linear model; UMAP, uniform manifold approximation and projection; NIfTI, Neuroimaging Informatics Technology Initiative.}
\label{fig:schematic}
\end{figure}

Here, we address these gaps with two objectives. First, we conduct the largest equity evaluation of open-source brain tumour segmentation models to date, assessing 18 architectures spanning the BraTS 2018--2023 challenge series across 648 patients from two independent datasets (UCSF-PDGM, $n = 501$; UPENN-GBM, $n = 147$)~\citep{calabrese2022ucsf, bakas2022upenn}. Our evaluation of these 11,664 model inferences operates across four equity dimensions: univariate group comparisons, multivariable regression, spatial meta-analysis, and high-dimensional representational equity analysis. Such an approach yields model-specific equity cards that provide a standardised fairness profile for each architecture. Second, we introduce Fairboard (\url{https://fairboard.streamlit.app}), an open-source, no-code Streamlit dashboard that implements all four equity dimensions, enabling researchers and clinicians to assess model fairness with zero programming expertise (\hyperref[fig:schematic]{Fig.~1}).

\section*{Results}

\subsection*{The Fairboard equitable model monitoring framework}

Fairboard is an open-source browser-based dashboard built in Streamlit that provides a unified interface for equitable model monitoring in neuroimaging (\hyperref[fig:schematic]{Fig.~1}). The platform accepts two input types: tabular performance data (individual-level or model-level metrics) and lesion segmentation masks (NIfTI volumes). Four analytical modules implement complementary equity dimensions: (i) \emph{Univariate Equity} provides non-parametric statistical tests, inequality metrics, and stratified visualisations; (ii) \emph{Cohort Equity} fits multivariable regressions to identify demographic predictors of model error; (iii) \emph{Spatial Equity} conducts multiple-comparison corrected voxel-wise generalised linear models (GLMs) to map any anatomically localised biases; and (iv) \emph{Representational Equity} performs dimensionality reduction with latent-space GLMs to detect if performance inequities exist within a nonlinear manifold, such as from an amalgamation of multi-modal image and text data about an individual. The outputs of Fairboard include interactive figures, downloadable PDF reports, exportable statistical tables, and NIfTI images. Fairboard supports five brain atlases, seven inequality metrics from the health economics literature~\citep{wagstaff2000measuring, gini1912variabilita, atkinson1970measurement, theil1967economics, hoover1936measurement, shorrocks1980class, palma2011homogeneous}, and both Bayesian and frequentist statistical frameworks. The dashboard is freely and openly available at \url{https://fairboard.streamlit.app}.

\subsection*{Qualitative segmentation variability across demographic groups}

\begin{figure}[H]
\centering
\includegraphics[width=\textwidth]{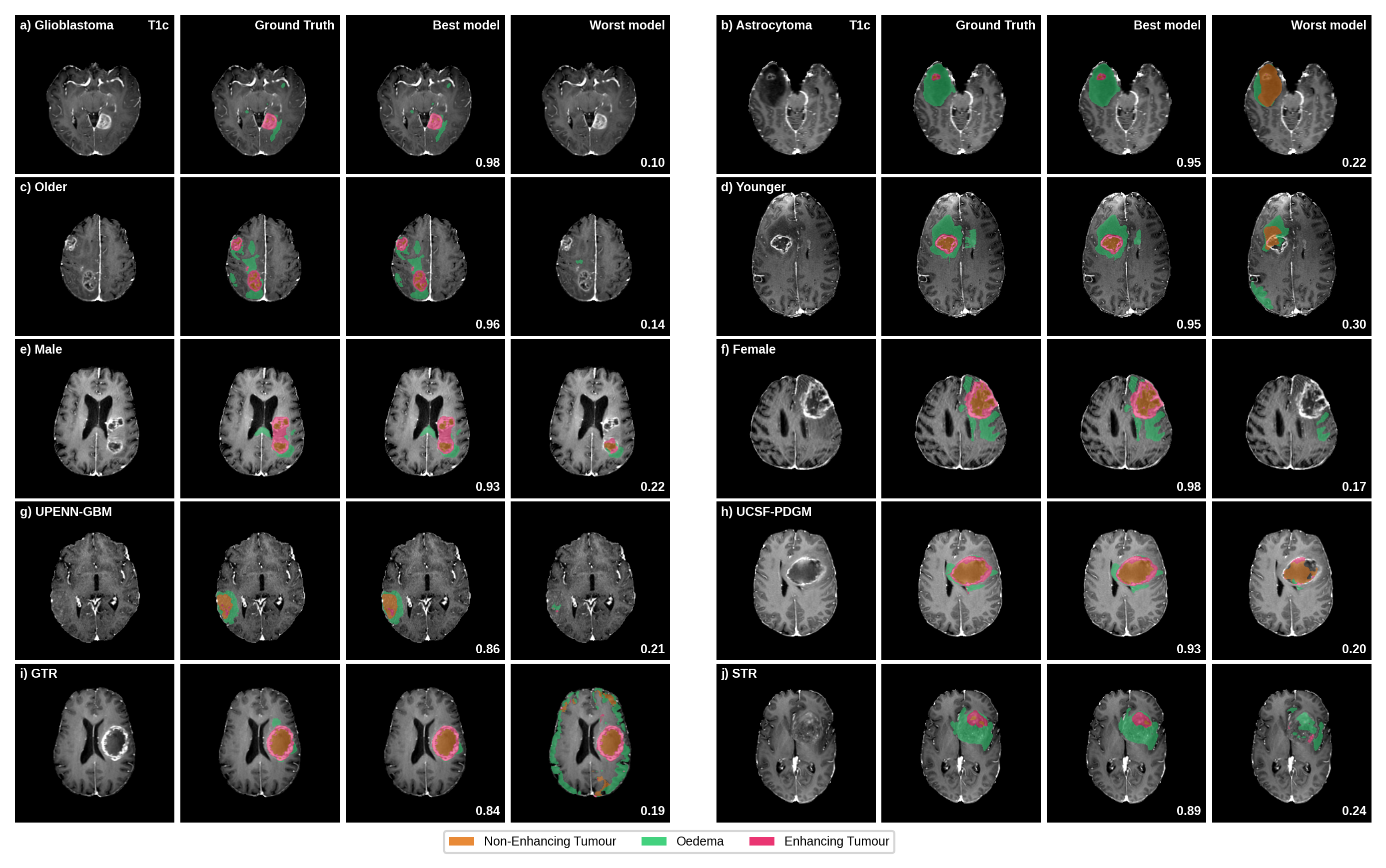}
\caption*{\textbf{Fig.~2. Qualitative segmentation comparison across diverse patient cases.}
Side-by-side comparison of sample brain tumour segmentations across all dimensions of available diversity in the datasets, illustrating variability in model performance. Patient cases are grouped by clinico-demographic factor: \textbf{(a--b)} molecular diagnosis (Glioblastoma, IDH-wildtype versus Astrocytoma, IDH-mutant), \textbf{(c--d)} age (older versus younger), \textbf{(e--f)} sex (male versus female), \textbf{(g--h)} dataset source (UPENN-GBM versus UCSF-PDGM), and \textbf{(i--j)} neurosurgical resectability (gross total resection versus subtotal resection). For each case, four columns are displayed (left to right): T1-weighted contrast-enhanced MRI scan, ground truth manual segmentation, best-performing model segmentation, and worst-performing model segmentation. Best and worst models were determined by mean DSC averaged across all four tumour compartments. Segmentation overlays: orange = non-enhancing tumour (NET), green = peritumoral oedema (OED), pink = enhancing tumour (ET). DSC, Dice similarity coefficient; GTR, gross total resection; STR, subtotal resection.}
\label{fig:qualitative}
\end{figure}

To motivate the quantitative analyses that follow, we first examined segmentation quality qualitatively across demographically diverse patient cases (\hyperref[fig:qualitative]{Fig.~2}). For each of five clinico-demographic factors (molecular diagnosis, age, sex, study site, and extent of surgical resection) we selected representative patient pairs and compared the best- and worst-performing models (ranked by mean Dice similarity coefficient [DSC] across all four tumour compartments). Visual inspection revealed substantial variability in segmentation quality both within and between models. These illustrative cases (each representing $n = 1$ per demographic stratum) are inherently anecdotal but motivate the systematic quantitative analyses presented in subsequent sections. An expanded comparison showing all 18 models is provided in Extended Data Fig.~1. We also make the inferences for all 648 patient cases from all 18 models ($n = 11{,}664$) available as supplementary data~\citep{ruffle2026fairboard_data}.

\subsection*{Performance and distributional equity are related, but not equivalent}

\begin{figure}[H]
\centering
\includegraphics[width=\textwidth]{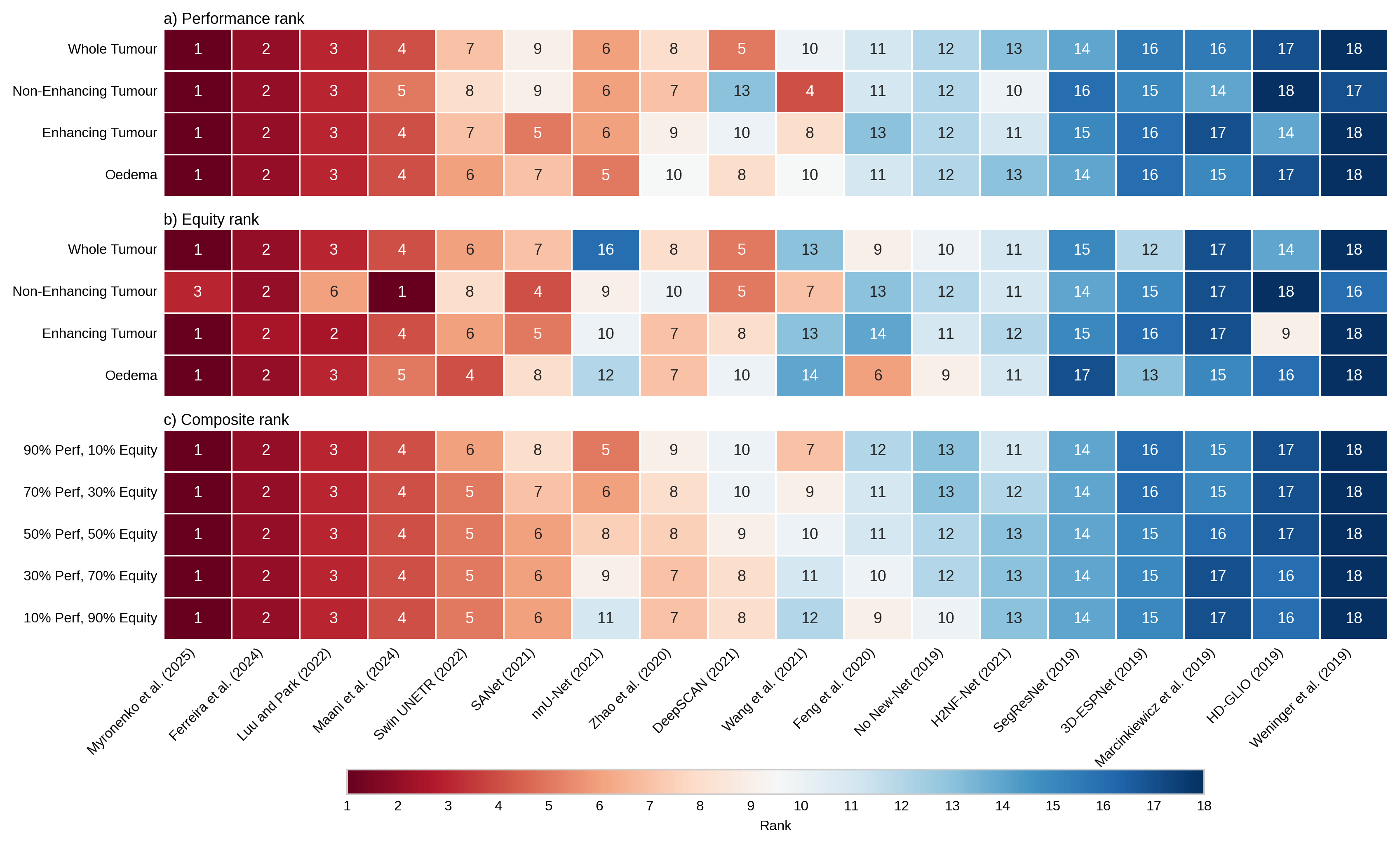}
\caption*{\textbf{Fig.~3. A composite league table of model performance and distributional equity.}
Compressed league table ranking 18 brain tumour segmentation models across three evaluation frameworks. \textbf{(a)} Performance rank, based on mean segmentation performance across seven metrics (DSC, sensitivity, precision, HD95, NSD@1mm, ASD, volume similarity) and four tumour compartments (WT, NET, ET, OED). \textbf{(b)} Equity rank, based on seven inequality metrics (Gini coefficient, Atkinson index, coefficient of variation, generalised entropy, Hoover index, Theil index, Palma ratio) computed across each of the 28 individual outcome measures (4 compartments $\times$ 7 metrics) at the patient level. \textbf{(c)} Composite rank under five weighting scenarios ranging from 90\% performance / 10\% equity to 10\% performance / 90\% equity. Colour scale: dark red = best rank (1), dark blue = worst rank (18). DSC, Dice similarity coefficient; HD95, 95th-percentile Hausdorff distance; NSD@1mm, normalised surface distance at 1\,mm tolerance; ASD, average surface distance; WT, whole tumour; NET, non-enhancing tumour; ET, enhancing tumour; OED, oedema.}
\label{fig:league}
\end{figure}

After excluding 72 patients with oedema-only ground-truth labels, 576 patients (429 UCSF-PDGM, 147 UPENN-GBM) were analysed here. We ranked all 18 models (Supplementary Table~1) simultaneously on segmentation performance and distributional equity to determine whether high-performing models are also distributionally equitable (\hyperref[fig:league]{Fig.~3}) and (Extended Data Fig.~2). Performance ranks were calculated from the mean scores across seven metrics (DSC, sensitivity, precision, 95th-percentile Hausdorff distance [HD95], normalised surface distance at 1\,mm [NSD@1mm], average surface distance [ASD], and volume similarity) and four tumour compartments (whole tumour [WT], non-enhancing tumour [NET], enhancing tumour [ET], and oedema [OED]). Equity ranks were derived from seven metrics of inequality in healthcare economics (Gini coefficient, Atkinson index, coefficient of variation, generalised entropy, Hoover index, Theil index and Palma ratio) calculated over the performance distribution at the patient-level for each model.

Composite rankings under five weighting scenarios (from 90\% performance / 10\% equity to 10\% performance / 90\% equity) revealed a broad trend: more recent, higher-performing models at the cohort level tend to be more distributionally equitable (\hyperref[fig:league]{Fig.~3c}) and (Extended Data Figs.~2--3). Myronenko et al.~\citep{myronenko2025brats23_2} (2nd place, BraTS 2023) ranked first on both performance and equity, with the three BraTS 2023 entrants~\citep{ferreira2024brats23_1,myronenko2025brats23_2,maani2024brats23_3} and the BraTS 2021 winner~\citep{luu2022kaist} occupying the top four positions across all weighting scenarios. Sensitivity--precision trade-off plots stratified by clinico-demographic group (Extended Data Fig.~4) and performance--equity scatter plots (Extended Data Figs.~5--11) confirmed a general positive association between higher segmentation accuracy and lower inequality across all seven equity metrics, all four tumour compartments, and the seven performance evaluation metrics, consistent with architectural advances, in general, having concurrently improved distributional fairness by virtue of minimising error rates in general. This relationship was however imperfect, with no model providing a formal distributional equity guarantee, given its performance. Univariate analyses revealed that models varied in their degree of demographic equity: for a given clinico-demographic factor, some models exhibited substantially larger subgroup performance gaps than others (Extended Data Figs.~12,~13). Age-stratified performance profiles showed some disparity across age brackets (Extended Data Fig.~14), and pairwise comparisons of all 153 model pairs (18 choose 2) confirmed significant inter-model differences in both performance and equity effect sizes (Extended Data Fig.~15).

\subsection*{Clinical and demographic factors predict segmentation performance}

\begin{figure}[H]
\centering
\includegraphics[width=\textwidth]{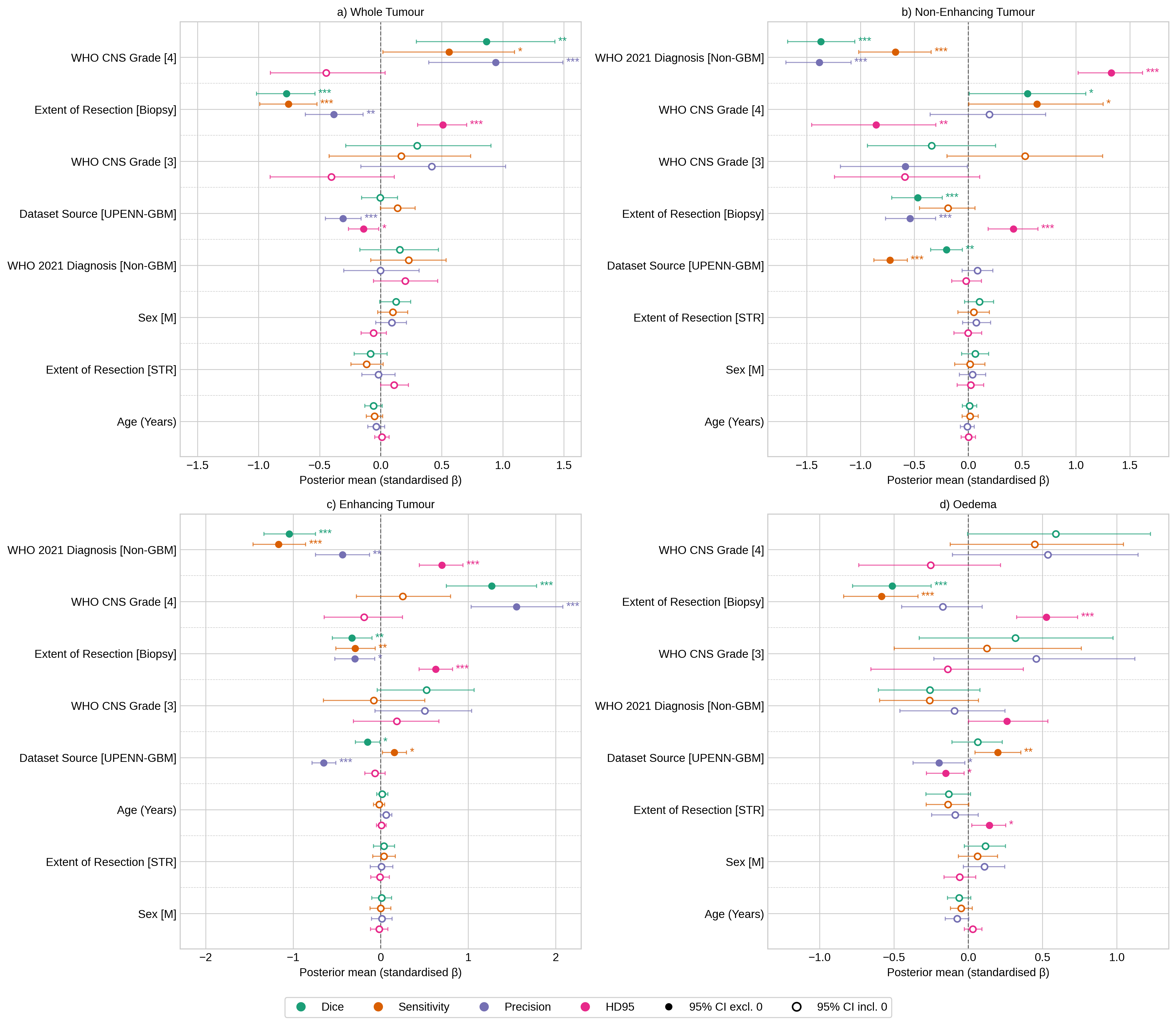}
\caption*{\textbf{Fig.~4. Bayesian linear mixed-effects model of clinico-demographic predictors of segmentation performance.}
Forest plots showing posterior mean standardised regression coefficients ($\beta$) and 95\% highest density intervals (HDI) from Bayesian linear mixed-effects (LME) models with crossed random intercepts for patient ($n = 569$) and model ($n = 18$). Fixed-effect predictor levels are: Sex [M], Age (Years), Dataset Source [UPENN-GBM], WHO CNS Grade [3], WHO CNS Grade [4], Extent of Resection [STR], Extent of Resection [Biopsy], and WHO 2021 Diagnosis [Non-GBM]. Panels show results for \textbf{(a)} whole tumour (WT), \textbf{(b)} non-enhancing tumour (NET), \textbf{(c)} enhancing tumour (ET), and \textbf{(d)} oedema (OED). Each predictor row displays four colour-coded points representing different performance metrics: teal = DSC, orange = sensitivity, purple = precision, magenta = HD95. Filled circles indicate effects where the 95\% HDI excludes zero; open circles indicate non-significant effects. Significance stars are based on the probability of direction (pd): $^{*}$pd $> 0.975$, $^{**}$pd $> 0.995$, $^{***}$pd $> 0.9995$. All continuous predictors and dependent variables were z-scored prior to model fitting. M, male; STR, subtotal resection; LME, linear mixed-effects; WT, whole tumour; NET, non-enhancing tumour; ET, enhancing tumour; OED, oedema; HDI, highest density interval; DSC, Dice similarity coefficient; HD95, 95th-percentile Hausdorff distance; pd, probability of direction.}
\label{fig:bayesian_lme}
\end{figure}

To identify which patient and model characteristics are associated with segmentation performance variation, we fitted Bayesian linear mixed-effects (LME) models with crossed random intercepts for patient ($n = 569$, after excluding $n = 7$ individuals with missing clinico-demographic data) and model ($n = 18$), yielding 9{,}165--10{,}242 observations per dependent variable depending on compartment-specific data availability (\hyperref[fig:bayesian_lme]{Fig.~4}). Six fixed-effect predictors were included: sex, age, dataset source, WHO Classification of Tumours of the Central Nervous System (CNS) grade, neurosurgical extent of resection, and WHO 2021 molecular diagnosis. All continuous predictors and dependent variables were z-scored prior to fitting. Models were estimated using Hamiltonian Monte Carlo (No-U-Turn Sampler; 4 chains $\times$ 6{,}000 draws, 3{,}000 warmup, target acceptance 0.95).

Variance decomposition revealed that patient identity consistently explained more variance than model identity. Patient-level intraclass correlation coefficients (ICCs) ranged from 0.31 (ET HD95) to 0.72 (NET Dice), whereas model-level ICCs ranged from 0.04 (NET sensitivity) to 0.22 (WT sensitivity), indicating that who the patient is, and the nature of their lesion, is associated with substantially more variance than which model is used. Conditional $R^2$ values (accounting for both fixed and random effects) ranged from 0.42 to 0.85. Marginal $R^2$ values (fixed effects only) ranged from 0.02 to 0.28, indicating that the six clinico-demographic predictors explained a modest fraction of total variance (Supplementary Table~2). Patient and model identity random intercepts accounted for a sizable proportion of explained variance, suggesting that unmeasured patient characteristics, lesional, or imaging properties may have also contributed to variations in segmentation performance that could be revealed in subsequent spatial or representational analyses.

Among the fixed effects, three clinical predictors exhibited the largest and most consistent effects across tumour compartments. \emph{Extent of resection} was the most consistently significant predictor: biopsy-only patients showed large performance deficits relative to gross total resection (GTR) (posterior mean $\beta$ ranging from $-0.77$ to $-0.33$ across Dice metrics; probability of direction [pd] $= 1.0$ for WT and NET Dice; pd $> 0.99$ for ET and OED Dice), while subtotal resection (STR) showed a trend toward reduced performance that reached significance only for HD95 oedema ($\beta = 0.14$, pd $= 0.99$). \emph{WHO 2021 molecular diagnosis} (non-Glioblastoma versus Glioblastoma, IDH-wildtype) was the strongest predictor for poorer NET and ET segmentation ($\beta = -1.37$ for NET Dice, $\beta = -1.05$ for ET Dice; both pd $= 1.0$). This seemed not unexpected, given the substantially different tumour morphologies of Glioblastoma, IDH-wildtype and the other gliomas, given the former often displaying greater degrees of post-contrast enhancement. \emph{WHO CNS grade} (grade 4 versus grade 2) showed strong positive effects on WT and ET metrics ($\beta = 0.87$ for WT Dice, $\beta = 1.27$ for ET Dice, $\beta = 1.55$ for ET precision; all pd $> 0.995$). Sex (male versus female) showed a smaller but detectable effect: male sex was associated with modestly higher WT Dice ($\beta = 0.13$, pd $= 0.97$). Age showed a weak negative trend for WT Dice ($\beta = -0.06$, pd $= 0.95$) that did not reach the pd $> 0.975$ significance threshold in any compartment. Full posterior coefficient estimates are provided in Supplementary Table~3. In summary, models generally performed better in higher grade and Glioblastoma, IDH-wildtype lesions that did not reach gross total resection at surgery, with male sex associated with modestly higher whole-tumour performance, and age showing small, inconsistent effects across compartments.

Prior sensitivity analysis across four candidate slope priors (bambi~\citep{capretto2022bambi} defaults, narrow Normal(0, 0.5), wide Normal(0, 5), and Cauchy(0, 1)) demonstrated robust inference, with differences in expected log pointwise predictive density (elpd$_{\text{WAIC}}$) below 2.0 for 15 of 16 dependent variables (maximum difference: 2.1; Supplementary Table~4). The best-fitting prior was selected per dependent variable via elpd$_{\text{WAIC}}$~\citep{vehtari2017practical}. Frequentist LME models with identical random-effects structure yielded concordant coefficient signs and significance patterns (Extended Data Fig.~16; Supplementary Table~5).

\subsection*{Voxel-wise meta-analysis identifies spatially localised segmentation bias}

\begin{figure}[H]
\centering
\includegraphics[width=\textwidth]{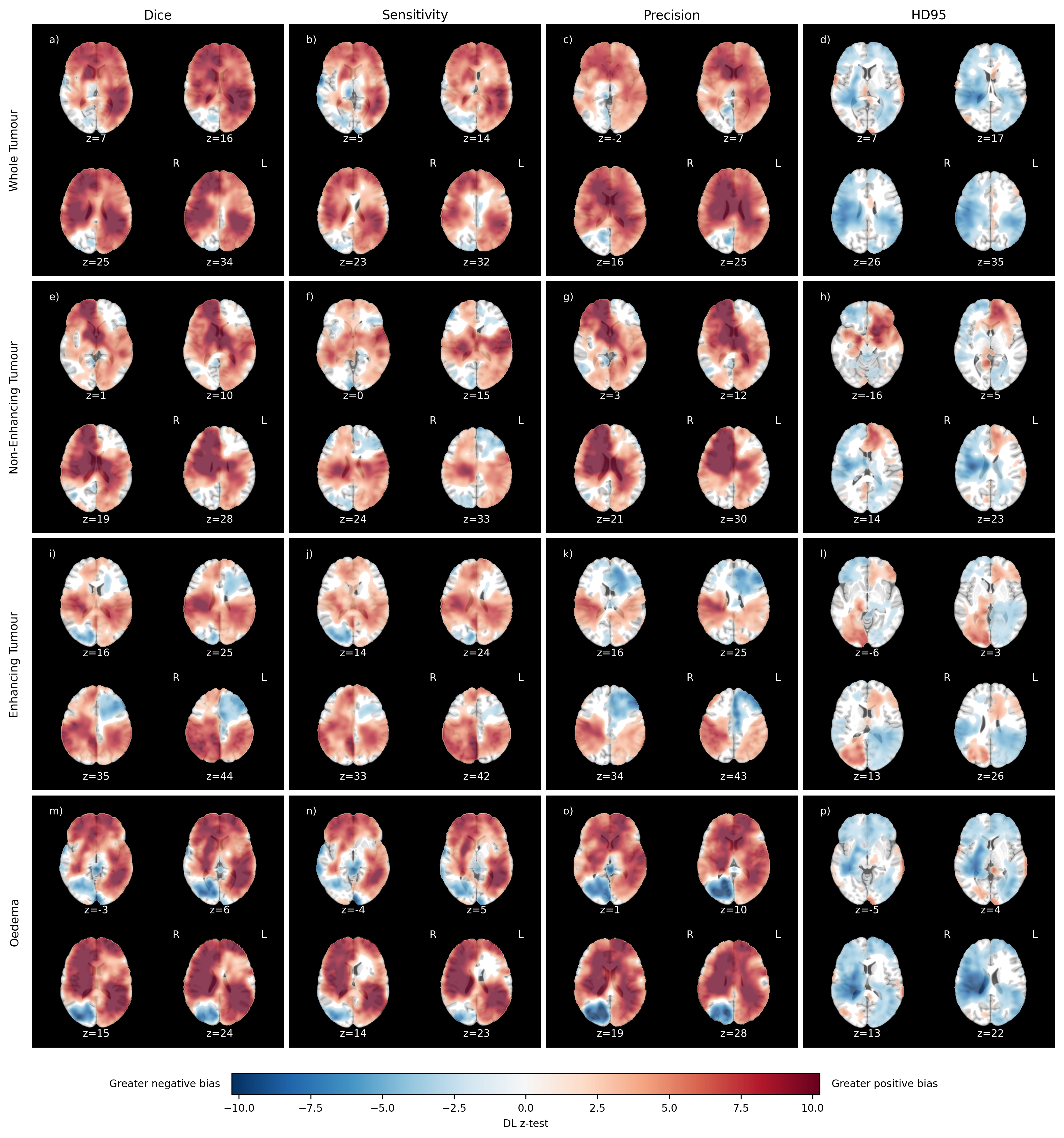}
\caption*{\textbf{Fig.~5. Spatial equity meta-analysis of voxel-wise segmentation bias.}
DerSimonian--Laird random-effects meta-analysis of voxel-wise segmentation performance bias across 18 models, Rows correspond to four tumour compartments (WT, NET, ET, OED) and columns correspond to four performance metrics (Dice, sensitivity, precision, HD95). Each cell shows multiple axial brain slices in MNI152 2\,mm standard space with voxel-wise combined z-statistics colour-mapped using a diverging colourmap (red/orange = positive bias indicating better performance; blue/cyan = negative bias indicating worse performance; note: for HD95 columns, this interpretation is reversed as higher HD95 values indicate worse boundary accuracy). The DerSimonian--Laird method estimates between-model heterogeneity ($\hat{\tau}^2$) and produces a pooled z-statistic at each voxel, accounting for inter-model variability. Voxels are thresholded using Benjamini--Hochberg FDR correction ($\alpha = 0.05$). DSC, Dice similarity coefficient; HD95, 95th-percentile Hausdorff distance; WT, whole tumour; NET, non-enhancing tumour; ET, enhancing tumour; OED, oedema; MNI, Montreal Neurological Institute; FDR, false discovery rate.}
\label{fig:spatial}
\end{figure}

We next asked whether segmentation biases manifest in spatially specific brain regions. For each of the 18 models, voxel-wise generalised linear models (GLMs) were fitted in Montreal Neurological Institute (MNI) 152 2\,mm standard space with age, sex, diagnosis, extent of resection, dataset source, and survival days as covariates. The resulting per-model z-statistic maps were then combined across models using a DerSimonian--Laird random-effects meta-analysis~\citep{dersimonian1986meta}, which estimates between-model heterogeneity ($\hat{\tau}^2$) and produces pooled z-statistics accounting for inter-model variability (\hyperref[fig:spatial]{Fig.~5}).

The meta-analytic z-maps revealed spatially structured biases across all four compartments and metrics (\hyperref[fig:spatial]{Fig.~5}). A sign-flipping permutation test (1{,}000 permutations) confirmed that observed spatial patterns far exceeded chance expectations: observed maximum $|z|$ ranged from 7.4 (ET HD95) to 19.2 (WT Dice), compared with 95th-percentile null maxima of 3.6--4.1 (family-wise error rate $p < 0.001$ for all eight compartment-metric combinations). For WT and OED, better Dice, Sensitivity, Precision and HD95 values were noted within occipital-sparing lesions, and also within the left hemisphere compared to the right. NET biases demonstrated that right anterior and bilateral basal ganglia/periventricular lesions yielded stronger segmentation performances, whereas for ET, better segmentation performance would be achieved in posterior frontal and parietal locations, with weaker anterior frontal performance, in general. Each tumour compartment thus exhibited distinct spatial patterns of performance bias, with the sign and magnitude of pooled z-statistics critically varying across brain regions (Supplementary Table~6). Lesion distribution heatmaps (Extended Data Fig.~17) provide anatomical context by showing population-level tumour prevalence across the brain, and vitally that the relationship between spatial performance was not simply a function of sampling size within a given locus. Signed model prevalence maps (Extended Data Fig.~18) quantified inter-model agreement on the direction of bias. We also provide per-model spatial equity cards with individual FDR correction (Extended Data Figs.~19--34), illustrating the model-specific patterns of regional strength and weakness. Voxel-wise $I^2$ heterogeneity was low across the majority of the brain (overall median $I^2 = 0$), indicating strong inter-model consistency; among voxels exhibiting non-zero heterogeneity (${\sim}8\%$ of brain voxels), the median $I^2$ was 0.47 (IQR 0.22--0.68), with WT Precision showing the highest between-model variability (median $I^2 = 0.86$) and NET Sensitivity the lowest (median $I^2 = 0.35$).

\subsection*{Representational equity reveals demographic clustering in latent performance space}

\begin{figure}[H]
\centering
\includegraphics[width=\textwidth]{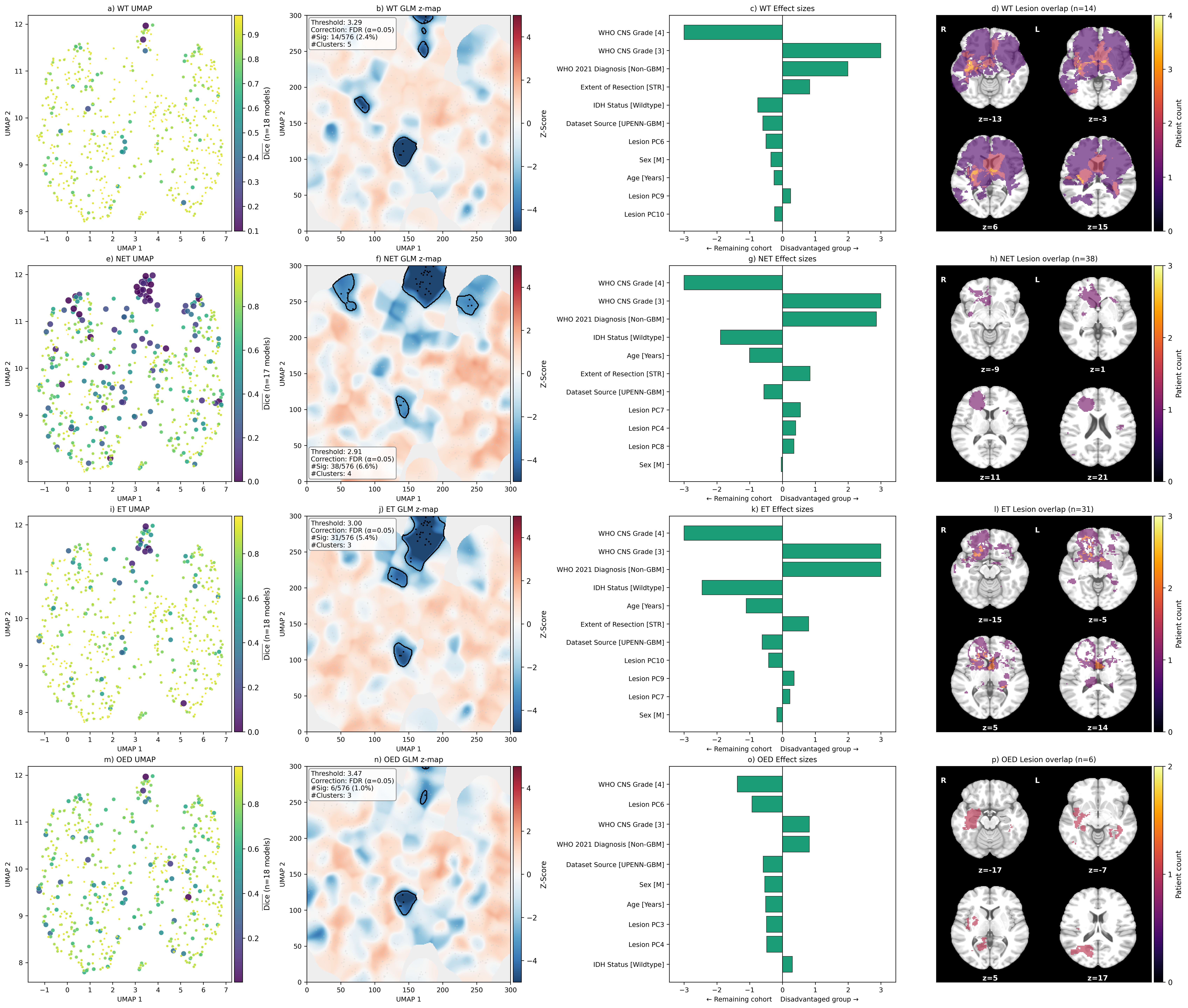}
\caption*{\textbf{Fig.~6. Representational equity analysis of Dice similarity across tumour compartments.}
High-dimensional representational equity analysis of DSC performance, shown separately for four tumour compartments (rows: WT, NET, ET, OED). Each row contains four panels. \textbf{Column~1:} UMAP scatter plot of individual patients ($n = 576$) in a latent space derived from PCA-compressed binary lesion mask features combined with encoded demographic variables. Points are coloured by mean DSC (yellow = high, purple = low). \textbf{Column~2:} Univariate GLM z-score map testing whether the z-scored compartment-specific DSC predicts structure in the UMAP latent space, with Benjamini--Hochberg FDR correction ($\alpha = 0.05$). \textbf{Column~3:} Cohen's $d$ effect sizes comparing significant versus non-significant patient clusters. \textbf{Column~4:} Lesion overlap heatmap in MNI152 standard space showing voxel-wise overlap of lesion masks from significant patients. UMAP, uniform manifold approximation and projection; PCA, principal component analysis; PC, principal component; GLM, generalised linear model; FDR, false discovery rate; MNI, Montreal Neurological Institute; DSC, Dice similarity coefficient; IDH, isocitrate dehydrogenase; WHO, World Health Organization; CNS, central nervous system; M, male; STR, subtotal resection.}
\label{fig:representational}
\end{figure}

Finally, to assess whether multi-modal patient data -- spanning imaging, clinical, and demographics -- systematically structure model performance in high dimensions, we conducted a representational equity analysis using Uniform Manifold Approximation and Projection (UMAP; \hyperref[fig:representational]{Fig.~6})~\citep{mcinnes2018umap}. A two-dimensional embedding was constructed from principal component analysis (PCA)-compressed multi-channel lesion mask features (10 principal components retaining 80\% of variance) combined with clinico-demographic variables age, sex, tumour grade, diagnosis, extent of resection, molecular IDH status, and dataset source. Segmentation performance metrics were excluded from the embedding to avoid circularity.

Latent-space GLMs tested whether z-scored model performance similarity, per lesional compartment, predicted spatial structure in the UMAP latent space, with Benjamini--Hochberg false discovery rate (FDR) correction at $\alpha = 0.05$~\citep{benjamini1995controlling}. Significant performance-related clustering was detected across all four tumour compartments, with the strongest effects in NET and ET (\hyperref[fig:representational]{Fig.~6}, columns 1--2). Cohen's $d$ effect sizes comparing patients in significant versus non-significant clusters revealed that the nonlinear amalgamation of patient demographics (age, sex), tumour histology (grade, molecular status and diagnosis), imaging appearance (lesion principal components), neurosurgical success (extent of resection), and geospatial features (dataset source) all contribute to model inequity within the manifold (\hyperref[fig:representational]{Fig.~6}, column 3). Lesion overlap maps of significant patients in MNI152 standard space showed that biased performance was also associated with specific anatomical distributions (\hyperref[fig:representational]{Fig.~6}, column 4), several of which reiterated the former spatial equity assessment. These findings were robust to UMAP hyperparameter choice: three alternative configurations (local: $n_{\text{neighbors}} = 5$; global: $n_{\text{neighbors}} = 30$; spread: $\text{min\_dist} = 0.3$) all detected significant performance-related clusters across all four compartments (3--8 clusters per configuration, max $|z| = 5.5$--$11.7$; Supplementary Table~7). Cohort-level representational equity analyses for HD95, precision, and sensitivity are presented in Extended Data Figs.~35--37, and per-model representational equity analyses in Extended Data Figs.~38--109. Together, the per-model spatial bias maps (Extended Data Figs.~19--34) and per-model representational equity analyses (Extended Data Figs.~38--109) constitute model-specific representational equity cards that provide a standardised summary of each architecture's fairness profile.

\section*{Discussion}

This study presents the most comprehensive equity evaluation of open-source brain tumour segmentation models to date, spanning 18 architectures, 648 patients from two independent datasets, and four complementary fairness dimensions. Four principal findings emerge. First, patient-level clinical characteristics (particularly extent of resection, WHO 2021 molecular diagnosis, and WHO CNS grade, alongside weaker effects of age and sex) are substantially stronger determinants of segmentation quality than model architecture, with patient-level ICCs consistently exceeding model-level ICCs by a factor of 2--19. Second, although more recent architectures, notably from BraTS 2023~\citep{ferreira2024brats23_1, myronenko2025brats23_2, maani2024brats23_3}, tend to rank higher on both performance and equity dimensions, no model yet provides formal equity guarantees, and explicit fairness-aware model selection remains necessary. Third, segmentation biases are not uniformly distributed across the brain but exhibit anatomically specific patterns that vary by tumour compartment and metric, as revealed by voxel-wise meta-analysis. Fourth, embedding multi-modal patient data---imaging morphology, clinico-demographic variables, and molecular markers---into a nonlinear latent space via UMAP reveals that segmentation inequity is not attributable to any single covariate in isolation but emerges from the joint, high-dimensional interaction of patient characteristics. 

These findings extend the medical AI fairness literature~\citep{mehrabi2022survey, mittermaier2023bias, ricci2022addressing}. Prior work has documented demographic biases in chest radiograph classification~\citep{seyyedkalantari2021underdiagnosis}, cardiac MRI segmentation~\citep{puyol2022fairness}, and medical imaging classification~\citep{larrazabal2020gender}, but these evaluations have largely relied on aggregate group comparisons. Our four-dimensional framework, combining univariate, multivariable, spatial, and representational analyses, shows that model inequity may vary across any of these approaches. Moreover, recent work has demonstrated that standard algorithmic fairness corrections can fail to generalise across clinical sites~\citep{yang2024limits} and that statistical fairness definitions do not necessarily translate into equitable health outcomes~\citep{stanley2025connecting}, underscoring the need for evaluation approaches that go beyond parity metrics alone. The use of inequality metrics from health economics~\citep{wagstaff2000measuring, gini1912variabilita, atkinson1970measurement, theil1967economics, hoover1936measurement, shorrocks1980class} offers a principled quantification of distributional equity that complements traditional group-level comparisons, but cannot disclose in greater detail the underlying source of inequity. The Bayesian LME approach applied here with crossed random effects enables simultaneous estimation of patient- and model-level variance components; in doing so, we identify that the dominant source of performance heterogeneity lies in patient characteristics rather than model choice. Models trained on narrow cohorts, such as in BraTS~\citep{menze2015brats, bakas2017advancing, baid2021rsna} with a focus on Glioblastoma, IDH-wildtype~\citep{louis2021who}, can skew model results. Such an approach will inevitably reflect the demographic composition of their training data, so weaker performance on younger patients, female patients, or non-glioblastoma cases is not an unexpected consequence~\citep{ruffle2023genetic}, though rarely considered.

These findings have several clinical implications. First, the performance-predicting effects of molecular diagnosis, WHO CNS grade, sex, age, study site, and neurosurgical extent of resection, have direct consequences for model deployment. Notably, the disparity related to extent of resection may reflect an underlying biological difference in tumour morphology across imaging and surgical vision. It seems plausible that for patients whose lesion is better delineated (and as such, AI-segmentable) on medical imaging, is also better delineated at operative time, and as such conveys a higher probability of achieving gross total resection. Conversely, tumours that are difficult to delineate algorithmically with imaging, may also be difficult to delineate surgically, implying a shared underlying challenge in boundary definition. Second, the effects of sex and age on segmentation quality, though smaller in magnitude than clinical predictors, indicate that demographic equity cannot be assumed even for characteristics often considered orthogonal to tumour biology. Male sex was associated with modestly higher whole-tumour segmentation performance, while age effects were small and inconsistent across compartments. Models should be validated across the full demographic spectrum, with particular attention to female patients, and the numerous other non-Glioblastoma, IDH-wildtype diagnoses, including but not limited to Astrocytoma, IDH-mutant, and Oligodendroglioma, IDH-mutant and 1p/19q co-deleted~\citep{louis2021who}. Third, the spatial localisation of biases revealed by the meta-analysis gives model developers concrete guidance: rather than seeking uniform improvements, targeted architectural modifications or training data augmentation could address region-specific weaknesses identified in the voxel-wise z-maps. At the clinical facing end, one should consider that the performance of any clinical decision-support tool is likely to vary depending on the neuroanatomical locus it is confined to. The observed laterality effect---where WT and OED segmentation performance was higher in the left hemisphere compared to the right---merits brief consideration. Possible explanations include laterality biases in training data composition, hemispheric asymmetries in tumour prevalence or morphology within glioma populations, or a chance finding.

The representational equity analysis warrants particular discussion, as it offers a fundamentally different lens on model inequity from the univariate, multivariable, and spatial approaches described above. Traditional fairness assessments stratify performance by individual demographic variables such as sex, age, and diagnosis, treating each as an independent axis of potential disparity~\citep{buolamwini2018gender, seyyedkalantari2021underdiagnosis}. Even multivariable regression, while accounting for confounders, assumes that covariates contribute additively (or at most through pre-specified interaction terms) to a linear predictor. The nonlinear latent embedding circumvents both limitations. By projecting the full multi-modal patient representation, encompassing imaging morphology (lesion mask principal components capturing shape, size, and compartment composition), clinico-demographic variables (age, sex, extent of resection), molecular markers (IDH mutation status, WHO 2021 diagnosis), tumour grade, and dataset provenance, into a unified two-dimensional manifold, UMAP captures complex, higher-order interactions among these features that no individual covariate or linear combination thereof can express~\citep{carruthers2022representational}. The resulting manifold geometry encodes patient similarity in a way that respects the nonlinear structure of the data: two patients who share the same sex, age bracket, and tumour grade may nonetheless occupy distant regions of the latent space because their lesion morphologies, molecular profiles, or resection histories differ in ways that jointly -- but not individually -- predict segmentation difficulty. Crucially, because segmentation performance was withheld from the embedding, the observation that model performance clusters significantly within the manifold is consistent with the patient feature space itself being intrinsically structured along axes of algorithmic vulnerability. This complements group-level comparisons: rather than showing that a demographic group receives worse performance on average, the representational analysis identifies the specific regions of the joint patient-characteristic space where models systematically underperform, and decomposes the contributions of demographic, molecular, morphological, and geospatial features to that underperformance via effect-size analysis~\citep{zemel2013learning, kearns2018preventing}. The clinical implication is that underserved patient populations cannot be adequately characterised by any single demographic label~\citep{crenshaw1989demarginalizing, lett2023intersectionality}. A patient who is female, young, harbours a low-grade Astrocytoma, IDH-mutant with diffuse non-enhancing morphology, and underwent subtotal resection occupies a compound region of the feature space that is poorly represented in standard Glioblastoma, IDH-wildtype dominated training cohorts. Yet, none of these individual characteristics in isolation would necessarily flag that patient as high-risk for segmentation failure~\citep{gichoya2022ai}. It is the nonlinear conjunction of these attributes that defines the patient's position within the fairness landscape, and it is this conjunction that the representational equity framework is uniquely positioned to detect~\citep{carruthers2022representational}.

Notably, several predictors (diagnosis, grade, extent of resection) contribute to both the LME and representational analyses, as expected---they are genuinely associated with performance. The complementarity underscores that the two approaches answer distinct questions: the LME identifies which individual covariates predict performance, while the UMAP analysis reveals whether the joint patient-characteristic space is structured along axes of model vulnerability. This approach may therefore serve as a general-purpose diagnostic for identifying underserved populations in medical AI, applicable beyond brain tumour segmentation to any setting where multi-modal patient data are available.

Several limitations warrant consideration. First, the Bayesian LME models included only random intercepts for patient and model; random slopes were not estimated due to computational constraints imposed by the crossed random-effects structure with $\sim$10{,}000 observations. If the effects of demographic predictors vary across models (e.g., if biopsy effects are larger for some architectures than others), the pooled fixed-effect uncertainty could be underestimated. However, a sensitivity analysis fitting random slopes for extent of resection (biopsy vs.\ non-biopsy) across the 18 models showed negligible shifts in all fixed-effect posterior means (max $|\Delta| = 0.009$ SD, mean $|\Delta| = 0.003$ SD; zero divergences), confirming that the pooled estimates are robust to this specification. The low model-level ICCs (0.04--0.22) further suggest that model identity contributes relatively little variance overall, which limits the scope for large random-slope variation. Second, the DerSimonian--Laird meta-analysis assumes independent studies, but the 18 models share test patients and, in some cases, architectural components (e.g., nnU-Net backbones), potentially inflating pooled z-statistic precision by underestimating correlated model errors. Comparing the precise architectural similarities and dissimilarities between models was beyond the scope of our research here, which instead was to prioritise the end result on diverse patients. This limitation is in-part mitigated by the complementary presentation of non-parametric median and prevalence maps, which do not assume study independence. Third, all performance metrics are computed against manual expert annotations, which are themselves subject to inter-rater variability (reported inter-rater Dice coefficients of $\sim$0.75--0.85 for enhancing tumour in the BraTS literature~\citep{menze2015brats}). If annotation quality varies systematically across lesion types, tumour grades, or institutions, this could confound performance comparisons across demographic groups. The UCSF-PDGM~\citep{calabrese2022ucsf} and UPENN-GBM~\citep{bakas2022upenn} datasets were annotated following standardised BraTS protocols, but we cannot exclude residual inter-institutional annotation variability. Fourth, the representational equity analysis relies on UMAP, whose stochastic nature means that embedding structure is partially dependent on hyperparameter choices; a supplementary UMAP hyperparameter sensitivity analysis across three alternative configurations (local: $n_{\text{neighbors}} = 5$, $\text{min\_dist} = 0.1$; global: $n_{\text{neighbors}} = 30$, $\text{min\_dist} = 0.1$; spread: $n_{\text{neighbors}} = 15$, $\text{min\_dist} = 0.3$) demonstrated that significant performance-related clusters were detected across all configurations and all four tumour compartments (3--8 clusters per configuration, max $|z| = 5.5$--$11.7$; Supplementary Table~7). Fifth, BraTS training set overlap was not modelled as a covariate in any analysis because the available flag we had acquired from the UCSF-PDGM and UPENN-GBM site leads tracks only BraTS 2021 training data, whereas the 18 evaluated models span challenges from 2018--2023. Including a partially valid confounder was judged worse than omitting it. Moreover, even where imaging data overlap exists, the paired clinical and demographic metadata used in our equity analyses were never provided as part of any BraTS challenge, so no model would have had the opportunity to calibrate for equitable performance across patient subgroups. Sixth, all estimates are derived from a GBM-dominant cohort (90.5\% Glioblastoma, IDH-wildtype; 95.3\% grade 4); non-GBM subgroups are small ($n = 55$), and the corresponding credible intervals are expectedly wider (e.g., WHO 2021 Diagnosis [Non-GBM] 95\% highest density interval width for NET Dice: 0.62; WHO CNS Grade 3: 1.19), reflecting limited statistical power for these subgroups.

A number of extensions would strengthen this work. Incorporating random slopes for key predictors in the LME framework would characterise the extent to which demographic effects vary across models. Expanding the evaluation to include models from the BraTS 2024 challenge and additional datasets (e.g., paediatric gliomas, metastatic disease) would test generalisability, and is a task for future research. Permutation-based inference for the representational equity GLMs would provide non-parametric validation of the current parametric findings. More broadly, the Fairboard dashboard is designed to be domain-agnostic: while this study focuses on brain tumour segmentation, the same four-dimensional equity framework could be applied to any healthcare modelling task where individual-level performance data are available~\citep{nikolov2021clinically, kather2022medical}. By providing an open-source, no-code tool for fairness assessment, Fairboard aims to make equity evaluation routine in medical AI development~\citep{chen2021ethical, obermeyer2019dissecting, petersen2023feature}.

In summary, this work establishes a four-dimensional equity framework for evaluating equity in healthcare AI models across univariate, multivariate, spatial, and representational planes. Across 18 brain tumour segmentation architectures and 648 patients, the dominant finding is that who a patient is matters more than which model segments their tumour: clinical and demographic characteristics consistently explain more performance variance than model architecture. The representational equity analysis further demonstrates that fairness cannot be fully understood through any single covariate or linear combination thereof, but instead requires characterisation of the joint, nonlinear patient feature space in which algorithmic vulnerability is embedded. Together with the release of our open-source Fairboard dashboard and model-specific equity cards for all 18 architectures, this framework provides both the conceptual and practical infrastructure for routine equity auditing in medical AI, offering a template that extends beyond neuro-oncology to any domain where equitable model deployment is a clinical imperative.

\section*{Methods}

This study was conducted in accordance with the Transparent Reporting of a multivariable prediction model for Individual Prognosis Or Diagnosis (TRIPOD) guidelines~\citep{collins2015tripod}, adapted for the model evaluation context.

\subsection*{Datasets}

Two publicly available glioma MRI datasets were used. The University of California San Francisco Preoperative Diffuse Glioma MRI dataset (UCSF-PDGM; $n = 501$ patients)~\citep{calabrese2022ucsf} provides preoperative multimodal MRI (T1-weighted, T1-weighted post-contrast, T2-weighted, and T2-fluid-attenuated inversion recovery [FLAIR] sequences) with expert manual segmentation labels for three tumour sub-regions: enhancing tumour, non-enhancing tumour and necrotic core, and peritumoral oedema~\citep{calabrese2022ucsf}. Accompanying clinical metadata include age, sex, WHO CNS grade, WHO 2021 molecular diagnosis, extent of resection, and survival status. The University of Pennsylvania Glioblastoma dataset (UPENN-GBM; $n = 147$ patients)~\citep{bakas2022upenn} provides analogous multimodal MRI with expert segmentations for patients with glioblastoma; clinical metadata include age, sex, extent of resection, and survival status. Whole tumour (WT) was defined as the union of the enhancing, nonenhancing tumour, and oedema sub-regions. No a priori power analysis was performed; sample size was determined by the availability of public datasets with expert segmentation labels and clinical metadata at the time of study.

We excluded patients whose ground-truth labels contained only an oedema label in the ground truth ($n = 72$). The remaining analysis cohort comprised 576 patients (429 UCSF-PDGM, 147 UPENN-GBM; 344 male, 232 female; mean age 60.0 $\pm$ 13.5 years, range 17--94). WHO 2021 molecular diagnoses~\citep{louis2021who} given in the source data~\citep{calabrese2022ucsf,bakas2022upenn} were Glioblastoma, IDH-wildtype in 521 patients (90.5\%), Astrocytoma, IDH-mutant in 44 (7.6\%), Astrocytoma, IDH-wildtype in 8 (1.4\%; reflecting legacy classification from the source dataset; under WHO 2021, these cases would typically be reclassified pending molecular marker assessment), and Oligodendroglioma, IDH-mutant and 1p/19q-codeleted in 3 (0.5\%). WHO CNS grades were distributed as grade 4 ($n = 549$, 95.3\%), grade 3 ($n = 17$, 3.0\%), and grade 2 ($n = 10$, 1.7\%). Extent of neurosurgical resection was gross total resection (GTR) in 335 patients (58.2\%), subtotal resection (STR) in 188 (32.6\%), and biopsy in 46 (8.0\%), with 7 patients (1.2\%) having missing resection data; regression analyses used the $n = 569$ patients with complete covariate data. Median overall survival was 384 days (interquartile range [IQR] 184--657; $n = 573$, 3 missing).

\subsection*{Brain tumour segmentation models}

\begin{sloppypar}
Eighteen publicly available brain tumour segmentation models spanning the BraTS 2018--2023 challenge series were evaluated (Supplementary Table~1). Three models were from BraTS 2023 (Ferreira et al.~\citep{ferreira2024brats23_1}, Myronenko et al.~\citep{myronenko2025brats23_2}, Maani et al.~\citep{maani2024brats23_3}), two from BraTS 2021 (Luu and Park~\citep{luu2022kaist}, Swin UNETR~\citep{hatamizadeh2022swinunetr}), five from BraTS 2020 (DeepSCAN~\citep{mckinley2021deepscan}, nnU-Net~\citep{isensee2021nnunet_brats2020}, SANet~\citep{yuan2021sanet}, Wang et al.~\citep{wang2021mpl}, H2NF-Net~\citep{jia2021h2nfnet}), one from BraTS 2019 (Zhao et al.~\citep{zhao2020bagoftricks}), six from BraTS 2018 (SegResNet~\citep{myronenko2019segresnet}, No New-Net~\citep{isensee2019nonewnet}, Feng et al.~\citep{feng2020ensemble3dunet}, Marcinkiewicz et al.~\citep{marcinkiewicz2019econib}, 3D-ESPNet~\citep{nuechterlein2019espnet}, Weninger et al.~\citep{weninger2019lfbrwth}), and one standalone model (HD-GLIO~\citep{kickingereder2019hdglio}, accessed via its public GitHub repository). BraTS 2018--2020 models were accessed via the BraTS Toolkit~\citep{kofler2020bratstoolkit}; SegResNet was accessed via the Medical Open Network for Artificial Intelligence (MONAI) framework. HD-GLIO is a two-class model that segments only ET and OED (not NET)~\citep{hdglio_github}; its NET metrics were set to missing (NaN) in all analyses.
\end{sloppypar}

\subsection*{Model inference and preprocessing}

All patient MRI volumes were preprocessed through the standardised BraTS pipeline, which includes co-registration of the four MRI sequences (T1, T1ce, T2, FLAIR), skull stripping, and nonlinear registration to the MNI152 template at 2\,mm isotropic resolution. Model inference was performed using published pretrained weights without retraining. Models compatible with CUDA compute capability 8.6 were run on an NVIDIA GeForce RTX 3090\,Ti GPU; older models requiring lower compute capability were run on an NVIDIA GeForce RTX 2080\,Ti GPU; models not supporting GPU acceleration were run with CPU fallback. All downstream equity analyses (Bayesian LME, voxel-wise GLMs, representational equity) were performed on CPU using the software stack described above.

\subsection*{Performance metrics}

Seven segmentation performance metrics were computed for each patient-model-compartment combination, following recommendations that no single metric suffices to characterise segmentation quality~\citep{maierhein2024metrics}. The Dice similarity coefficient (DSC)~\citep{dice1945measures} quantifies volumetric overlap between predicted and ground-truth segmentations. Sensitivity (recall) and precision measure the fraction of true positives detected and the fraction of positive predictions that are correct, respectively. The 95th-percentile Hausdorff distance (HD95)~\citep{huttenlocher1993comparing} measures the worst-case boundary error after excluding the most extreme 5\% of surface distances. The normalised surface distance at 1\,mm tolerance (NSD@1mm) quantifies the fraction of the predicted surface within 1\,mm of the ground-truth surface. Average surface distance (ASD) measures the mean bidirectional surface distance. Volume similarity quantifies the ratio of predicted to ground-truth volumes. All metrics were computed for four tumour compartments: WT, NET, ET, and OED, yielding 28 performance measures per patient-model pair.

\subsection*{Inequality metrics}

Seven inequality metrics from the health economics literature~\citep{wagstaff2000measuring, regidor2004measures} were adapted to quantify distributional equity of model performance across patients. The Gini coefficient~\citep{gini1912variabilita} measures the area between the Lorenz curve and the line of perfect equality, ranging from 0 (perfect equality) to 1 (maximal inequality). The Atkinson index ($\varepsilon = 0.5$)~\citep{atkinson1970measurement} measures the fraction of total performance that could be sacrificed without reducing social welfare if performance were equally distributed, where the inequality aversion parameter $\varepsilon$ controls sensitivity to the lower tail. The normalised coefficient of variation (CoV) is computed as $\text{cv} / (\text{cv} + 1)$, where $\text{cv}$ is the ratio of standard deviation to mean, bounding the index to $[0, 1)$. The generalised entropy index ($\alpha = 2$)~\citep{shorrocks1980class} belongs to a parametric family controlling sensitivity to different parts of the distribution, with $\alpha = 2$ emphasising the upper tail. The Hoover index (Robin Hood index)~\citep{hoover1936measurement} represents the proportion of total performance requiring redistribution to achieve equality. The Theil index~\citep{theil1967economics}, a member of the generalised entropy family ($\alpha = 1$), is fully decomposable into within-group and between-group components. The Palma ratio~\citep{palma2011homogeneous} measures the ratio of the income share of the top 10\% to the bottom 40\%, capturing inequality at the distribution extremes. All metrics were computed over the patient-level performance distribution for each model, requiring that the distribution minimum be shifted to $10^{-6}$ for metrics that are undefined for zero-valued inputs.

\subsection*{Univariate equity assessment}

Univariate analyses assessed whether segmentation performance differed across demographic strata for each model independently. Equity gap bars were computed as the difference in mean performance between two demographic subgroups per model, with 95\% bootstrap confidence intervals (1,000 iterations, percentile method). For naturally binary stratifications (sex, dataset source), groups were compared directly. Multi-level demographics were binarised for gap computation: WHO CNS grade (grade 4 versus non-grade 4), extent of resection (GTR versus STR), and WHO 2021 diagnosis (Glioblastoma, IDH-wildtype versus non-glioblastoma (Oligodendroglioma, IDH-mutant and 1p/19q-codeleted or Astrocytoma, IDH-mutant)). Age-stratified performance was visualised across fine-grained age bins ($<$30, 30--39, 40--49, 50--59, 60--69, 70--79, 80+ years) to assess non-linear trends without imposing parametric assumptions.

\subsection*{Bayesian linear mixed-effects models}

To quantify the joint contribution of demographic predictors to segmentation performance, Bayesian linear mixed-effects (LME) models were fitted with crossed random intercepts for patient ($n = 569$) and model ($n = 18$) using bambi~\citep{capretto2022bambi} and PyMC~\citep{salvatier2016pymc3}. Six fixed-effect predictors were included: sex, age (z-scored), dataset source (reference: UCSF-PDGM), WHO CNS grade (reference: grade 2), extent of neurosurgical resection (reference: gross total resection), and WHO 2021 molecular diagnosis (reference: Glioblastoma, IDH-wildtype). Each of the 16 dependent variables (4 compartments $\times$ 4 metrics: DSC, sensitivity, precision, HD95) was modelled separately. The model formula was:
\begin{equation}
y \sim x_1 + x_2 + \ldots + x_p + (1 \mid \text{patient}) + (1 \mid \text{model})
\end{equation}
where $y$ is a z-scored performance metric, $x_1 \ldots x_p$ are the $p = 8$ dummy-coded predictor columns derived from the six conceptual predictors (sex, age, dataset source, two grade dummies, two resection dummies, and diagnosis), and the two random intercept terms capture patient- and model-level variance. The Gaussian likelihood was used with an identity link. Variance decomposition, including $R^2$ marginal (fixed effects only) and $R^2$ conditional (fixed plus random effects), was computed from the posterior mean variance components following Nakagawa and Schielzeth~\citep{nakagawa2013general}. Intraclass correlation coefficients were calculated as $\text{ICC}_j = \sigma^2_j / (\sigma^2_{\text{model}} + \sigma^2_{\text{patient}} + \sigma^2_{\text{residual}})$ for each random-effects grouping variable $j$.

Hamiltonian Monte Carlo sampling was performed using the No-U-Turn Sampler (NUTS) with 4 chains, 6,000 posterior draws per chain, 3,000 warmup iterations, and a target acceptance probability of 0.95. Convergence was assessed via the potential scale reduction factor ($\hat{R}$) and bulk effective sample size (ESS)~\citep{vehtari2017practical}. Across all 16 dependent variables, $\hat{R} \leq 1.01$ and minimum bulk ESS ranged from 319 (NET Dice) to 3{,}795 (ET HD95), with 15 of 16 dependent variables exceeding an ESS of 500. The NET Dice model yielded the lowest minimum ESS (319), below the commonly recommended threshold of 400; posterior estimates for this variable should therefore be interpreted with additional caution. No divergent transitions were observed for any model. All sampling used a fixed seed with \texttt{random\_state} = 42 for reproducibility.

Prior sensitivity was evaluated by fitting each dependent variable under four candidate slope prior specifications: bambi defaults (data-driven), narrow Normal(0, 0.5), wide Normal(0, 5), and Cauchy(0, 1). Group-level random effects used Normal priors with HalfNormal hyperpriors for the variance parameter in all configurations. The residual standard deviation used a HalfStudentT prior. The best-fitting prior was selected per dependent variable via expected log pointwise predictive density from the Widely Applicable Information Criterion (elpd$_{\text{WAIC}}$). Posterior inference was based on 95\% highest density intervals (HDIs), with effects considered significant when the HDI excluded zero (equivalent to pd $> 0.975$). Significance stars were assigned based on the probability of direction (pd): $^{*}$pd $> 0.975$, $^{**}$pd $> 0.995$, $^{***}$pd $> 0.9995$, corresponding approximately to two-sided frequentist $p < 0.05$, $p < 0.01$, and $p < 0.001$ respectively. As a robustness check, model comparison was additionally performed using leave-one-out cross-validation (elpd$_{\text{LOO}}$) via ArviZ~\citep{kumar2019arviz}; rankings were concordant with elpd$_{\text{WAIC}}$ selections.

\subsection*{Frequentist regression models}

As a concordance check, frequentist LME models with identical fixed-effect predictors and crossed random-effects structure were estimated via restricted maximum likelihood (REML) using statsmodels~\citep{seabold2010statsmodels}. Statistical significance of fixed-effect coefficients was assessed at $\alpha = 0.05$ with Benjamini--Hochberg FDR correction~\citep{benjamini1995controlling} across the 16 dependent variables. Multiple testing correction was applied independently within each analysis domain (16 Bayesian LME models, 16 frequentist LME models, 16 spatial meta-analyses, 4 representational GLMs) using Benjamini--Hochberg FDR at $\alpha = 0.05$. This domain-stratified approach reflects the exploratory, hypothesis-generating nature of each equity dimension, where the four analytical frameworks address distinct scientific questions and operate on different data representations.

\subsection*{Spatial equity analysis}

For each of the 18 models, voxel-wise GLMs were fitted in MNI152 2\,mm standard space using nilearn's \texttt{SecondLevelModel}~\citep{abraham2014nilearn}. Lesion masks, disaggregated by compartment (WT, ET, NET, OED), were smoothed with an 8\,mm full-width-at-half-maximum (FWHM) Gaussian kernel, consistent with the default smoothing kernel in voxel-based morphometry~\citep{ashburner2000vbm}, prior to GLM fitting. Smoothed binarised lesion masks served as the dependent variable, with the patient-level z-scored segmentation performance metric as the predictor of interest alongside nuisance covariates (age, sex, diagnosis, extent of resection, dataset source, and survival days). The resulting per-model z-statistic maps identified voxels where lesion presence was significantly associated with segmentation performance after covariate adjustment; these were individually thresholded using Benjamini--Hochberg FDR correction at $\alpha = 0.05$~\citep{benjamini1995controlling} (Extended Data Figs.~19--34).

These 18 per-model z-maps were then combined using the DerSimonian--Laird random-effects meta-analysis~\citep{dersimonian1986meta}. For each voxel, the Cochran $Q$ statistic was computed as $Q = \sum_{i=1}^{k}(z_i - \bar{z})^2$ (assuming equal within-study variance of 1 for z-scores), the between-study variance estimated as $\hat{\tau}^2 = \max(0, (Q - (k-1)) / (k-1))$, and the pooled z-statistic computed as $z_{\text{DL}} = \bar{z} / \sqrt{(1 + \hat{\tau}^2) / k}$, where $k = 18$ (or 17 for NET metrics, excluding HD-GLIO). The $I^2$ heterogeneity statistic~\citep{higgins2002quantifying} was computed as $I^2 = \max(0, (Q - (k-1)) / Q) \times 100\%$. Voxels were thresholded using Benjamini--Hochberg FDR correction at $\alpha = 0.05$~\citep{benjamini1995controlling}, applied separately per compartment-metric combination. The equal unit-variance assumption is a simplification; if per-model GLMs differ in effective sample size or goodness of fit, within-study variances will vary, potentially biasing the pooled estimate toward models with larger true variance. We further note that the DerSimonian--Laird method assumes independent studies; because all models are evaluated on the same patient set, pooled z-statistics may be anti-conservative. A complementary non-parametric summary (signed prevalence map) is therefore provided in Extended Data Fig.~18. To validate the meta-analytic findings under the violated independence assumption, a sign-flipping permutation test was conducted: for each of 1{,}000 permutations, each model's z-map was randomly multiplied by $\pm 1$, a permuted DL z-map was recomputed, and the maximum $|z|$ was recorded to form a null distribution. Observed maximum $|z|$ values (7.4--19.2 across eight compartment-metric combinations) exceeded the 95th-percentile null maxima (3.6--4.1) by a factor of 1.8--5.3$\times$ (family-wise error rate $p < 0.001$ for all eight tests), confirming that the spatial biases are not artefactual.

\subsection*{Representational equity analysis}

Representational equity analysis assessed whether patient demographics predict structure in a latent space constructed from lesion morphology and clinical features. For each patient, binary lesion masks from the ground-truth segmentation were resampled to a common isotropic $64 \times 64 \times 64$ voxel grid and flattened, then compressed via PCA (retaining $K = 15$ components explaining 80\% of variance). These lesion features were concatenated with one-hot encoded demographic variables (sex, WHO CNS grade, extent of resection, WHO 2021 diagnosis, dataset source, IDH status) and z-scored continuous variables (age). All features were standardised to zero mean and unit variance prior to UMAP embedding.

\begin{sloppypar}
UMAP~\citep{mcinnes2018umap} was applied to the combined feature matrix with standard default hyperparameters ($n_{\text{neighbors}} = 15$, $\text{min\_dist} = 0.1$) and cosine distance metric, producing a two-dimensional embedding (\texttt{random\_state} = 42 for reproducibility). Segmentation performance metrics were explicitly excluded from the embedding to prevent circularity.
\end{sloppypar}

The two-dimensional UMAP coordinates were then rasterised onto a $300 \times 300$ grid, where each patient's position was represented as a Gaussian-smoothed spike (FWHM = 18 grid units, corresponding to 6\% of the grid extent; chosen to balance spatial resolution with adequate coverage for the sample size, analogous in proportional extent to the $\sim$4\% spanned by the standard 8\,mm FWHM kernel in voxel-based morphometry at 2\,mm isotropic resolution~\citep{ashburner2000vbm}). This image representation enabled the use of nilearn's \texttt{SecondLevelModel} to test whether z-scored performance metrics (e.g., compartment-specific DSC) predicted spatial structure in the latent space, with Benjamini--Hochberg FDR correction at $\alpha = 0.05$. A coverage mask excluded grid cells containing no actual patient data. Cohen's $d$ effect sizes were computed comparing patients falling within significant clusters to those outside, across all demographic variables and the top 15 lesion principal components. This analysis was conducted at the cohort level (averaging performance across all 18 models) and separately for each model (Extended Data Figs.~38--109).

\subsection*{Fairboard dashboard}

The Fairboard dashboard was implemented in Python~3.10 using the Streamlit framework (v1.48). For representational equity and latent-space analyses, users may select from four dimensionality reduction algorithms: principal component analysis (PCA; via scikit-learn~\citep{pedregosa2011scikitlearn}), $t$-distributed stochastic neighbour embedding ($t$-SNE~\citep{vandermaaten2008tsne}; configurable perplexity 5--50, default 30), uniform manifold approximation and projection (UMAP~\citep{mcinnes2018umap}; configurable $n_{\text{neighbors}}$ 2--50, default 15), and non-negative matrix factorisation (NMF~\citep{lee1999nmf}). Brain parcellation analyses support six atlas options fetched via nilearn~\citep{abraham2014nilearn}: Schaefer 2018~\citep{schaefer2018atlas} (100--1,000 parcels, 7 or 17 Yeo networks), Harvard--Oxford cortical and subcortical atlases (distributed with FSL~\citep{jenkinson2012fsl}), the Automated Anatomical Labeling (AAL) atlas~\citep{tzourio2002aal}, the J\"{u}lich probabilistic cytoarchitectonic atlas~\citep{eickhoff2005juelich}, and the Destrieux 2009 sulcogyral atlas~\citep{destrieux2010atlas}, each available at 1 or 2\,mm resolution.

Bayesian regression models use bambi~\citep{capretto2022bambi} (v0.15) with PyMC~\citep{salvatier2016pymc3} (v5.25), exposing user-configurable Markov chain Monte Carlo (MCMC) sampling parameters: posterior draws per chain (500--10,000; default 4,000), number of chains (1--8; default 4), CPU cores for parallel sampling, and target acceptance rate (0.70--0.99; default 0.90), with 1,000 warm-up draws. Bayesian model diagnostics use ArviZ~\citep{kumar2019arviz} (v0.23). Frequentist regression is available via ordinary least squares (OLS) through statsmodels~\citep{seabold2010statsmodels} (v0.14.2), with optional feature normalisation. Distributional equity metrics include the Gini coefficient, Atkinson index, Theil index, Hoover index, Palma ratio, generalised entropy index, and normalised coefficient of variation, implemented as custom Python modules within the Fairboard codebase. Statistical testing parameters are configurable throughout, including significance threshold (default $\alpha = 0.05$), post-hoc correction method (Benjamini--Hochberg FDR, Bonferroni, or uncorrected), and optional test/train splitting. Spatial equity analyses allow adjustment of voxel resolution (2--8\,mm), smoothing kernel FWHM (0--16\,mm), and significance threshold. Cloud deployment is detected automatically, with Bayesian sampling parameters adjusted accordingly. Thread-safe session state management ensures reliable operation under concurrent use.

\subsection*{Data availability}

\begin{sloppypar}
The UCSF-PDGM dataset is publicly available at The Cancer Imaging Archive (\url{https://www.cancerimagingarchive.net/collection/ucsf-pdgm/}).
The UPENN-GBM dataset is publicly available at The Cancer Imaging Archive (\url{https://www.cancerimagingarchive.net/collection/upenn-gbm/}).
All model inferences (11,664 segmentation volumes and per-patient performance metrics) are publicly available on Zenodo~\citep{ruffle2026fairboard_data}.
\end{sloppypar}

\subsection*{Software availability}

The Fairboard dashboard is freely available at \url{https://fairboard.streamlit.app}.

\subsection*{Ethics statement}

This study used exclusively publicly available, de-identified datasets (UCSF-PDGM~\citep{calabrese2022ucsf} and UPENN-GBM~\citep{bakas2022upenn}) obtained from The Cancer Imaging Archive. Both datasets were collected under institutional review board approval at their respective institutions and made publicly available with appropriate de-identification. No additional ethics approval was required for secondary analysis of these public datasets. No new data were collected from human participants.

\subsection*{Author contributions}

J.K.R.\ conceived the study, developed the Fairboard dashboard, performed all analyses, and wrote the manuscript. S.M., C.F., M.Z.\ and Z.W.\ undertook user testing and debugging of analyses and Fairboard, and reviewed the manuscript. S.B.\ and H.H.\ provided data acquisition and reviewed the manuscript. P.N.\ conceived the study, provided infrastructural support, and reviewed the manuscript.

\subsection*{Competing interests}

The authors declare no competing interests.

\subsection*{Funding}

This work was supported by the British Society of Neuroradiology (J.K.R.), the Medical Research Council (UKRI1389 \& MR/X00046X/1) (J.K.R.), the Wellcome Trust (P.N.), and the UCLH NIHR Biomedical Research Centre (J.K.R., H.H., P.N.).

\bibliographystyle{unsrtnat}

\end{document}